\documentclass[a4paper,12pt]{spieman}  
\usepackage{amsmath,amsfonts,amssymb}
\usepackage{graphicx}
\usepackage{setspace}
\usepackage{tocloft}

\newcommand{\elem}{\,\in\,}
\newcommand{\bn}[1]{\makebox[3ex][c]{#1}}
\newcommand{\nnshape}[3]{$ \bn{#1} \times \bn{#2} \times \bn{#3} $}
\newcommand{\convk}[2]{$ #1 \times #2 $}
\newcommand{\convks}[3]{$ #1 \times #2 \;/\; #3 $}
\newcommand{\kk}{\convk{3}{3}}
\newcommand{\kks}{\convks{3}{3}{2}}
\newcommand{\ktrans}{$K^{\mathit{trans}}$}

\title{Deep Generative Adversarial Neural Networks for Realistic Prostate Lesion MRI Synthesis}

\author[a]{Andy Kitchen}
\author[b]{Jarrel Seah}
\affil[a,*]{Independent Researcher}
\affil[b]{STAT Innovations Pty. Ltd., PO Box 274, Ashburton VIC 3147, Australia}

\cftpagenumbersoff{figure}
\cftpagenumbersoff{table} 
\begin{document} 
\maketitle

\begin{abstract}

	Generative Adversarial Neural Networks (GANs) are applied to the
	synthetic generation of prostate lesion MRI images. GANs have been
	applied to a variety of natural images, is shown show that the same
	techniques can be used in the medical domain to create realistic
	looking synthetic lesion images. $16\text{mm} \times 16\text{mm}$
	patches are extracted from 330 MRI scans from the SPIE ProstateX
	Challenge 2016 and used to train a Deep Convolutional Generative
	Adversarial Neural Network (DCGAN) utilizing cutting edge techniques.
	Synthetic outputs are compared to real images and the implicit latent
	representations induced by the GAN are explored. Training techniques
	and successful neural network architectures are explained in detail.

\end{abstract}

\keywords{Deep Learning, Generative Adversarial Networks, MRI, Convolutional Neural Network}

{\noindent \footnotesize\textbf{*}Andy Kitchen,
	\linkable{contact@andy.kitchen} }

\begin{spacing}{1}

\section{Introduction}

	Generative Adversarial Neural Networks (GANs) are state of the art
	machine learning models that can learn the statistical regularities of
	input data and then generate a nearly endless stream of synthetic
	examples that resemble, but do no exactly replicate, the input
	data\cite{NIPS2014_5423}.  These models have been applied to generate a
	variety of natural images, including images of bedrooms,
	faces\cite{DBLP:journals/corr/RadfordMC15} and
	animals\cite{NIPS2016_6125}.  In this work GANs are applied to
	generate realistic looking synthetic images of prostate lesions
	resembling the SPIE ProstateX Challenge 2016 training data.  Multiple
	aligned MRI modalities are generated simultaneously, and the model
	produces compelling results with a relatively small amount of training
	data.

	The ability to create synthetic data that resembles real data in key
	statistical aspects is well studied and particularly important in the
	medical field where anonymity is critical. In the appropriate
	circumstances, machine learning or data mining can be carried out on
	surrogate synthetic data instead of raw sensitive data, giving improved
	anonymization. When there are only a small number of training examples,
	generated data can be used as extra training data, a powerful way to
	combat overfitting and increase model performance\cite{NIPS2016_6125}.
	Synthetic image generation can also be used as an aid for education and
	medical training.

\section{Generative Models}

	Generative models are distinct from discriminative models because they
	capture the distribution of data itself, instead of the conditional
	probability of a label given data. This data distribution can then be
	sampled; a process of generating new data that `looks like' real data.

	\begin{align*}
		P(Y|X) &\qquad \textbf{Discriminative Model} \\
		P(X)   &\qquad \textbf{Generative Model}
	\end{align*}

	\noindent Where $X$ is a random variable taking values from the input
	domain and $Y$ is a random variable of associated labels. $P(Y|X)$ is
	the conditional distribution of labels given data and $P(X)$ is the
	data distribution itself.

	Generative models are often much more complicated than discriminative
	models. They must capture all the intricacies of data not just the
	parts specific to a label. For example, with a blood test, a
	statistical model may only need to look for elevated levels in one or
	two dimensions to indicate a pathology, but generating a whole new
	blood panel that looks as if it had come from a patient would require
	capturing complex interdependencies between different levels.

\section{Generative Adversarial Models}

	An adversarial model is formalized as a game played between two
	players, with distinct competing objectives. Called the generator $G$
	and a discriminator $D$. $G$ is an `artist' that tries to create
	realistic looking images. $D$ is a `critic' that tries to classify
	images as fake, created by the artist; or as a real images sampled from
	the world. The principal equilibrium strategy in this game is for $G$
	to draw from $P(X)$ in which case $D$ performs no better than random
	guessing, i.e. the best way for $G$ to fool $D$ is to create images
	that are indistinguishable from real images (according to $D$).

	Interestingly, unlike most models, there is no global loss function that
	must be minimized, instead these models are trained to an equilibrium
	point where neither player can improve their performance given a small
	unilateral change to their strategy; where their strategy is
	represented by continuous neural network weights. A leap-frog gradient
	descent algorithm is used for training, where a gradient descent step
	is taken for $G$ with $D$ held constant, then $D$ with $G$ held
	constant.  With some luck and under conditions that are in general not
	well understood this algorithm can move both players into a suitable
	equilibrium strategy.
	
	This method is particularly powerful if the discriminative models are
	large Deep Convolutional Neural Networks. If there are any recognizable
	statistical aberrations in the data generated by $G$ then $D$ can
	catch out the generator by recognizing these aberrations. Unrealistic
	structures are thus suppressed when training has reached equilibrium
	--- $G$ produces highly realistic samples.

\section{Practical Training of GANs}

	GANs are already notorious for being hard to train, equilibrium
	strategies are often unstable and hard to reach compared to the optima
	of a single function. If either $G$ or $D$ are too powerful, one will
	dominate the other, gradients will vanish and the models will become
	stuck in a poor equilibrium, often producing images that look like
	noise or have no content. In general $G$ and $D$ must be designed
	together and matched in terms of power i.e. they should be commensurate
	in terms of layer size and depth.  Implementors should be aware that
	only certain combinations of generator and discriminator will work well
	together, and compatibility is hard to predict in advance. The authors
	recommend iterative development informed by existing literature,
	intuition, and empirical testing.

	It can also be beneficial to introduce a large amount of activation
	noise and dropout into $D$, allowing $G$ to compete with a wide variety
	of slightly different strategies; this can help to escape from poor
	equilibria. Using batch normalization and special activation functions
	has also show to be effective in some
	cases\cite{DBLP:journals/corr/RadfordMC15}.

\section{Method}
\subsection{Data Preparation}

\begin{figure}
\begin{center}
  \includegraphics[]{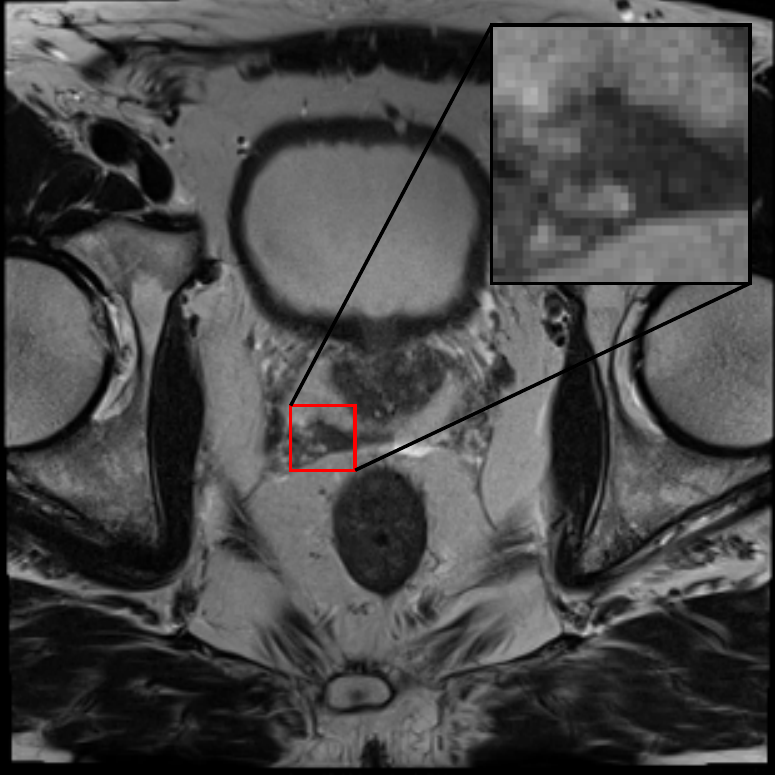}
\end{center}
\caption{T2-weighted MRI with Patch Region}
\label{fig:patch}
\end{figure}

	All training data is extracted from the SPIE ProstateX Challenge 2016
	data set and prepared using the same method the authors used for
	competition entries\cite{kitchen2017}.  Patches of $16\text{mm} \times
	16\text{mm}$ in size are extracted around the centres of 330 prostate
	lesion MRI scans at a resolution of $1\text{px}/\text{mm}$. Three
	modalities are aligned and utilized: T2, ADC and \ktrans{}.
	All channels are normalized to approximately lie in the range $[0, 1]$.
	Each input image patch has three channels, one for each modality. See
	figure \ref{fig:patch} for diagram.

\subsection{Generator Architecture}

\begin{figure}[p]
\begin{center}
  \includegraphics[]{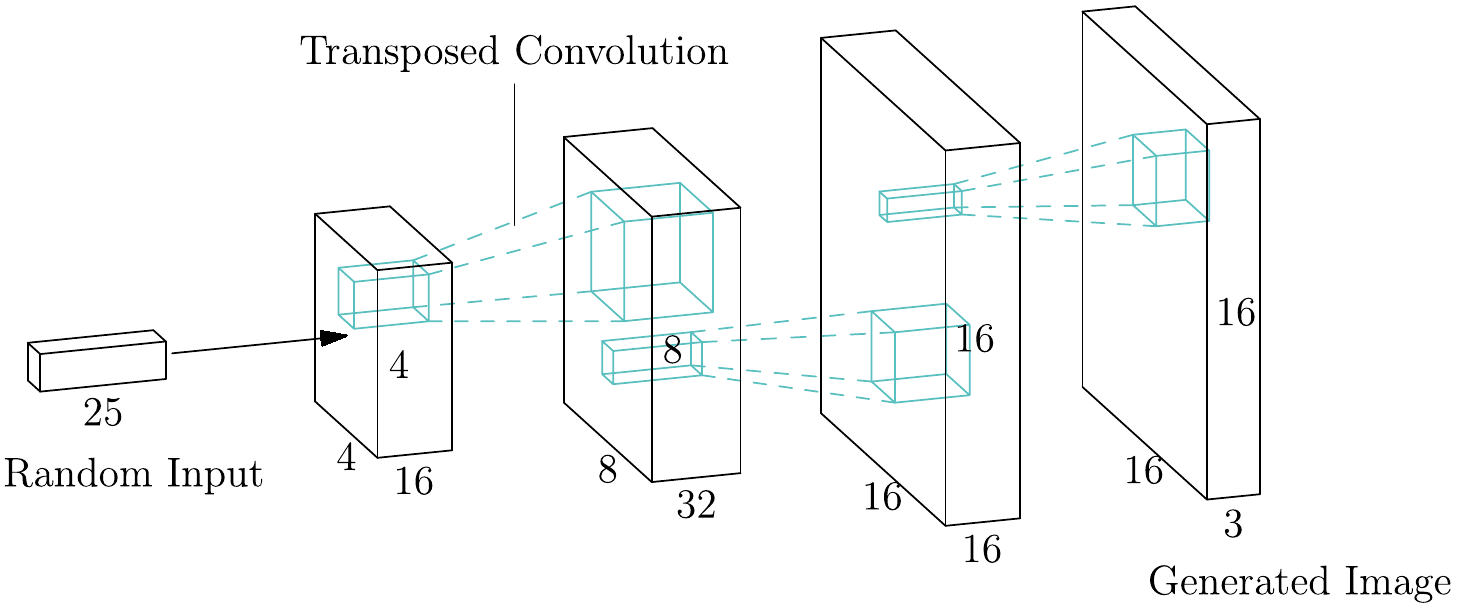}
\end{center}
\caption{Generator Neural Network Schematic}
\label{fig:generator}
\end{figure}

\begin{table}[h]
\begin{center}
  \begin{tabular}{c | l | r | c}
    & kernel & feats. & out. shape \\
    \hline
    random  input    &         &       & $25$                 \\
    fully connected  &         & $265$ & $256$                \\
    reshape          &         &       & \nnshape{4}{4}{16}   \\
    T. conv. / ReLU  & \kks    &  $32$ & \nnshape{8}{8}{32}   \\
    T. conv. / ReLU  & \kks    &  $16$ & \nnshape{16}{16}{16} \\
    T. conv. / ReLU  & \kks    &   $3$ & \nnshape{16}{16}{3}  \\
  \end{tabular}
\end{center}
\caption{Generator Neural Network Details}
\label{tab:generator}
\end{table}

	\noindent The generator neural network has 5 layers and includes
	transposed convolutional layers \cite{Dumoulin2016} (also called
	`deconvolutional' layers). The input is a 25 dimensional vector of
	standard normal random numbers, followed by a fully connected layer and
	3 transposed convolutions.  See figure \ref{fig:generator} for a
	schematic and table \ref{tab:generator} for layer details.

\subsection{Discriminator Architecture}

\begin{figure}[p]
\begin{center}
  \includegraphics[]{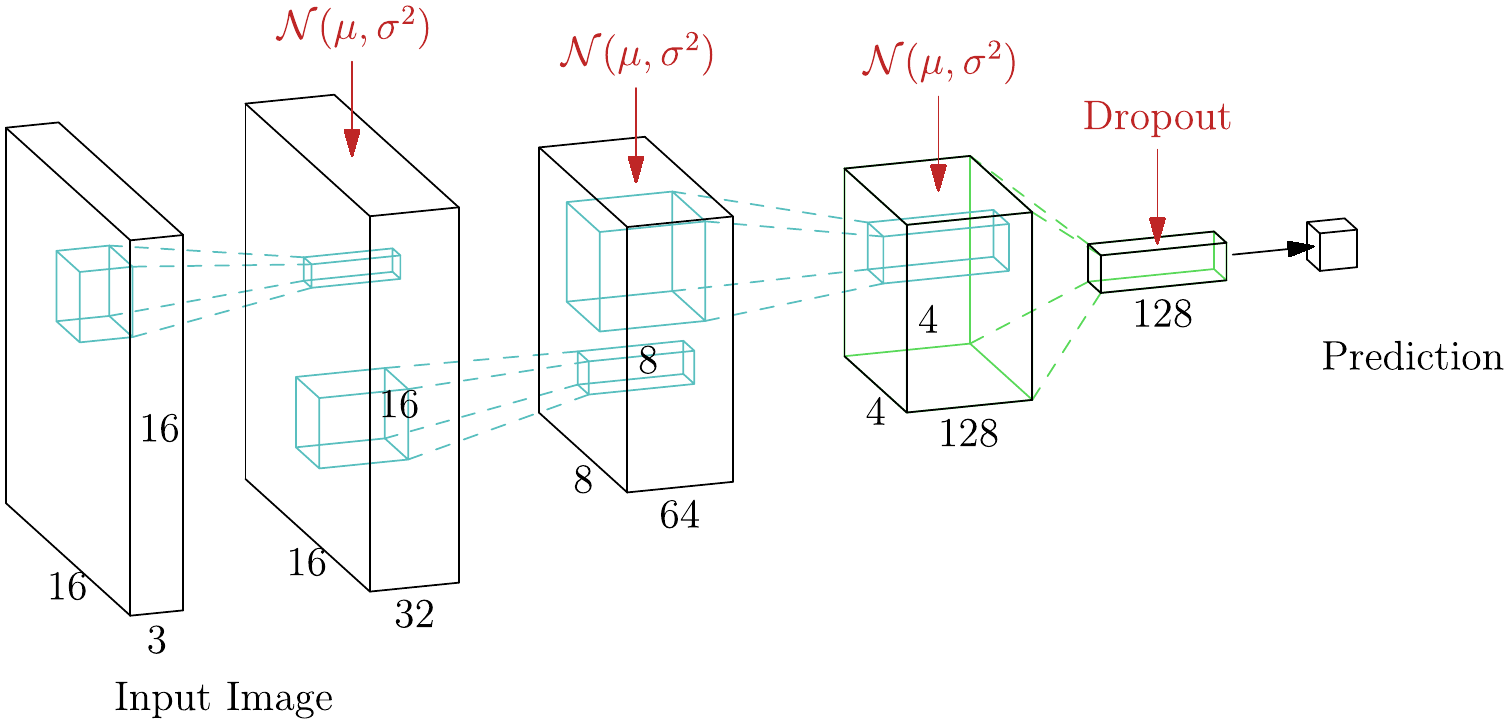}
\end{center}
\caption{Discriminator Neural Network Schematic}
\label{fig:discriminator}
\end{figure}

\begin{table}[h]
\begin{center}
  \begin{tabular}{c | l | r | c | l}
    & kernel & feats. & out. shape & noise \\
    \hline
    input            &         &     & \nnshape{16}{16}{3}  & gaussian \\
    conv. / L. ReLU  & \kk     & 32  & \nnshape{16}{16}{32} & gaussian \\
    conv. / L. ReLU  & \kks    & 64  & \nnshape{8}{8}{64}   & gaussian \\
    conv. / L. ReLU  & \kks    & 128 & \nnshape{4}{4}{128}  & gaussian \\
    global avg. pool &         &     & \nnshape{1}{1}{128}  & dropout  \\
    fully connected  &         &     & $1$                  &          \\
  \end{tabular}
\end{center}
\caption{Discriminator Neural Network Details}
\label{tab:discriminator}
\end{table}
	
	\noindent The discriminator neural network has 6 layers, an initial
	image input layer, 3 layers of convolutions followed by global average
	pooling and a fully connected layer. The final hidden layer uses
	dropout; all other hidden layers have gaussian noise added. To try and
	improve gradient flow by preventing saturation `leaky' ReLU activation
	functions are used: $\max(\alpha x, x)$, where $\alpha = 0.1$ in this
	work. Gaussian noise is drawn from $\mathcal{N}(0, 1/2)$. See figure
	\ref{fig:discriminator} for a schematic and table
	\ref{tab:discriminator} for details.

\subsection{Training Objective}

	Formulas essentially the same as the empirical cross entropy are used
	for both $G$ and $D$ loss functions\cite{NIPS2014_5423}:

	\begin{align}
		\label{eqn:ldloss}
		L_D(\theta_G, \theta_D) &=
		-\; \frac{1}{n}
		\sum_{x \elem G(\theta_G, z)} \log(1-p(x; \theta_D))
		-
		\frac{1}{m}
		\sum_{x \elem \bar{X}} \log(p(x; \theta_D))
		\\
		L_G(\theta_G, \theta_D) &=
		\phantom{-}\; \frac{1}{n}
		\sum_{x \elem G(\theta_G, z)} \log(1-p(x; \theta_D))
	\end{align}

	\noindent Where $L_D$ is the discriminator loss function, $L_G$ is the
	generator loss function. $\theta_G$ and $\theta_D$ are the respective
	neural network parameters. $p(x; \theta_D)$ is the probability that $D$
	assigns to $x$ being real.  $G(\theta_G,z)$ is a set of images
	generated by $G$ for random normal inputs $z$. $\bar{X}$ is a sample of
	natural images from $P(X)$. The first sum of equation \ref{eqn:ldloss}
	is taken over fake images and penalizes high probabilities from $D$, the
	second term is taken over real images and penalizes low probabilities.
	$n$ is the number of fake images in a batch, $m$ is the number of real
	images in a batch.

\subsection{Training Procedure}

	A leapfrog gradient descent is used to find an equilibrium point of the
	GAN game. The following updates are iterated until convergence:

	\begin{align}
		\theta_G^{i+1} &\xleftarrow{\text{Adam}}
		\frac{\partial L_G(\theta_G^i, \theta_D^i)}
		     {\partial \theta_G^i}
		\\
		\theta_D^{i+1} &\xleftarrow{\text{Adam}}
		\frac{\partial L_D(\theta_G^i, \theta_D^i)}
		     {\partial \theta_D^i}
	\end{align}

	\noindent Where $\theta_G$ is a vector of generator neural network
	parameters, and $\theta_D$ are the discriminator parameters, $L_G$ and
	$L_D$ are their respective loss functions. The arrow indicates
	application of the Adam accelerated gradient descent algorithm for the
	update\cite{DBLP:journals/corr/KingmaB14}. The model is trained for
	15,000 iterations with a batch size of 200 (200 fake and 200 real
	images).

\section{Results}

\begin{figure}
\begin{center}
  \includegraphics[]{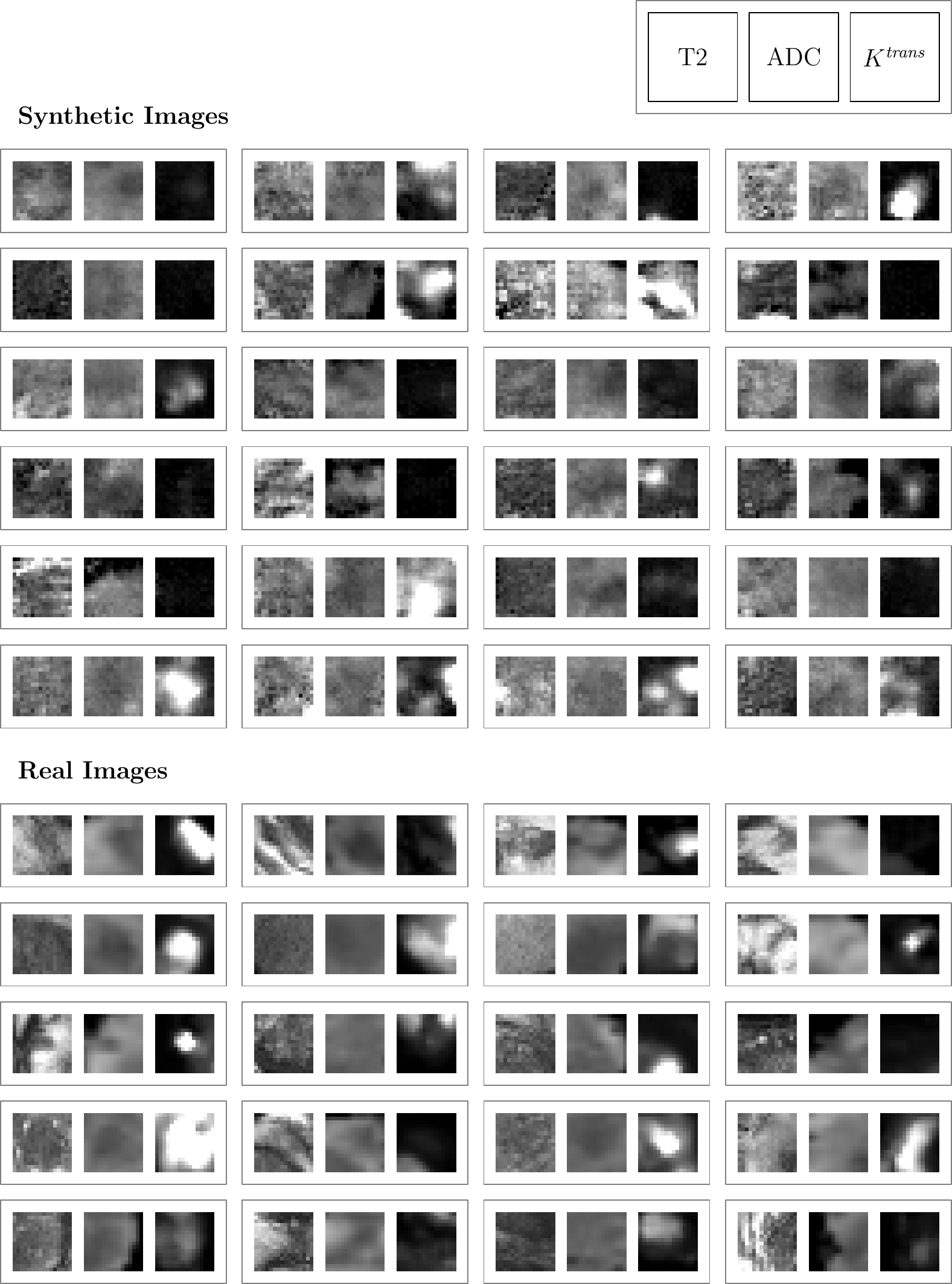}
\end{center}
\caption{Comparison of Real and Synthetic Image Patches}
\label{fig:synthetic}
\end{figure}

\begin{figure}
\begin{center}
  \includegraphics[]{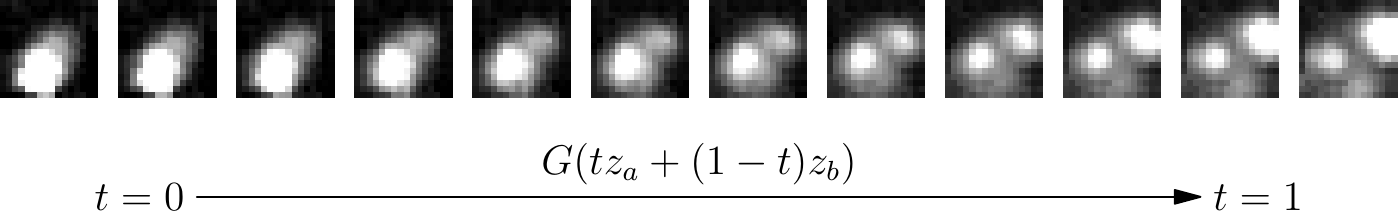}
\end{center}
\caption{Interpolation in $Z$ space}
\label{fig:interpolation}
\end{figure}

	See figure \ref{fig:synthetic} for a full page comparison of real and
	synthetic images. Qualitatively the synthetic T2 mode has captured
	the rough broken textures of the real patches, the ADC mode correctly
	darkens the lesion centre. The \ktrans{} mode displays large coherent
	blobs similar to how they appear in real data, notice that bright
	\ktrans{} areas are accompanied by matching darker regions in the ADC
	mode, this is a benefit of simultaneously generating all modes
	together, they are coherent with each other.

	For any random input $z$, $G(z)$ should fool $D$ with a high
	probability. Thus the input space $Z$ of $G$ forms an implicit latent
	representation of prostate lesions. See figure \ref{fig:interpolation}
	for example of linear interpolation between two lesion images in $Z$
	space.  There is a smooth transition between two lesion morphologies,
	demonstrating the high quality of the implicit latent representation.

\section{Disclosures}

	All included research has been independently self funded by the authors
	outside of the institutional system.  No conflicts of interest,
	financial or otherwise, are declared by the authors.

\section{Acknowledgments}

	The authors would like to acknowledge the organizers of the SPIE
	ProstateX Challenge 2016 for their hard work in organizing the
	competition and preparing the training data used in this work.

\bibliography{article}
\bibliographystyle{spiejour}

\vspace{1\baselineskip}

\noindent
\begin{minipage}[c]{0.2\textwidth}
\includegraphics[width=0.75\textwidth]{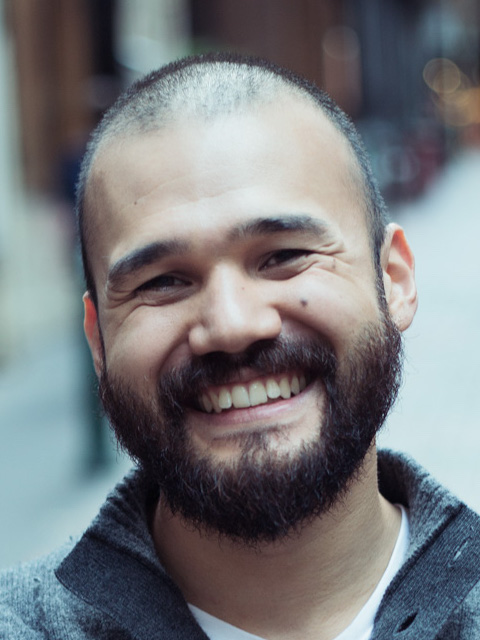}
\end{minipage}
\begin{minipage}[c]{0.8\textwidth}

	\noindent\textbf{Andy Kitchen} is a Machine Learning and AI researcher
	with experience in Deep Learning and Neural Networks. He has served as
	lead engineer and founder of two startup companies.  During the last
	two years he has headed a Data Science and Machine Learning consulting
	team. Andy has received multiple data science awards, including first
	prizes for teams in the Australian HealthHack and GovHack competitions.
	He is currently applying for PhD positions.

\end{minipage}

\vspace{1\baselineskip}

\noindent
\begin{minipage}[c]{0.2\textwidth}
\includegraphics[width=0.75\textwidth]{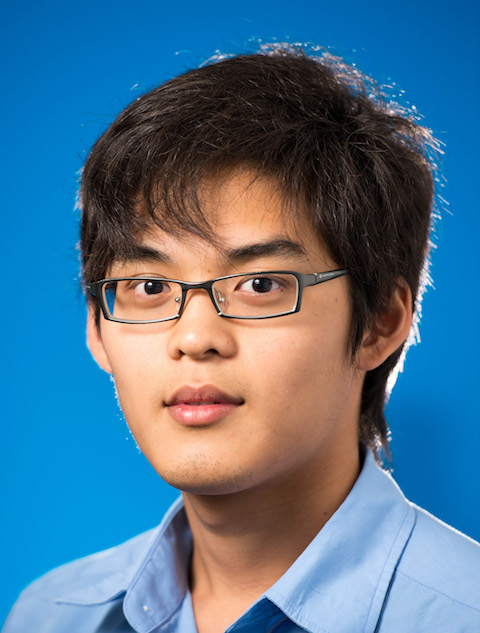}
\end{minipage}
\begin{minipage}[c]{0.8\textwidth}

	\noindent\textbf{Jarrel Seah} is a doctor, researcher and
	cofounder of STAT innovations. He is a winner of the prestigious
	Microsoft Imagine Cup for the Eyenemia app. He believes that Machine
	Learning, Artificial Intelligence, Virtual Reality will fundamentally
	change health care.

\end{minipage}

\vspace{1\baselineskip}

\listoffigures
\listoftables

\end{spacing}
\end{document}